\documentclass[final]{cvpr}

\usepackage{times}
\usepackage{epsfig}
\usepackage{graphicx}
\usepackage{bbm}
\usepackage{caption}
\usepackage[font=small]{caption}
\usepackage{amsmath}
\usepackage{amssymb}
\usepackage{enumitem}
\usepackage{relsize}
\usepackage{cite}   
\usepackage{subfloat}
\usepackage{tabularx}
\usepackage{adjustbox}
\usepackage{multirow}
\usepackage[flushleft]{threeparttable}
\usepackage{tablefootnote}
\usepackage{lipsum}

\usepackage{tabulary}
\usepackage{booktabs}       
\usepackage{amsfonts}       
\usepackage{nicefrac}       
\usepackage{microtype}      
\usepackage{subfig}

\newcommand{\tabfref}[1]{Table~\ref{#1}}
\newcommand{\eqnref}[1]{Eq.~(\ref{#1})}
\renewcommand{\paragraph}[1]{\vspace{1mm}\noindent\textbf{#1}}

\usepackage{xcolor}
\definecolor{cadmiumorange}{rgb}{0.93, 0.53, 0.18}

\usepackage[pagebackref=true,breaklinks=true,colorlinks,bookmarks=false]{hyperref}

\begin{document}

\title{Learning to Associate Every Segment for Video Panoptic Segmentation}

\author{Sanghyun Woo\textsuperscript{1} \quad 
Dahun Kim\textsuperscript{1} \quad
Joon-Young Lee\textsuperscript{2} \quad
In So Kweon\textsuperscript{1}\\ \\
\textsuperscript{1}KAIST \qquad \textsuperscript{2}Adobe Research
}
\maketitle
\pagestyle{empty}
\thispagestyle{empty}

\begin{abstract}
Temporal correspondence - linking pixels or objects across frames - is a fundamental supervisory signal for the video models.
For the panoptic understanding of dynamic scenes, we further extend this concept to every segment.
Specifically, we aim to learn coarse \textbf{segment-level} matching and fine \textbf{pixel-level} matching together.
We implement this idea by designing two novel learning objectives.
To validate our proposals, we adopt a deep siamese model and train the model to learn the temporal correspondence on two different levels (i.e., segment and pixel) along with the target task. At inference time, the model processes each frame independently without any extra computation and post-processing.
We show that our per-frame inference model can achieve new state-of-the-art results on Cityscapes-VPS and VIPER datasets. 
Moreover, due to its high efficiency, the model runs in a fraction of time (3$\times$) compared to the previous state-of-the-art approach.
\end{abstract}

\section{Introduction}

A holistic understanding of a video requires pixel-level information of different semantics, object instances, background stuffs as well as their temporal changes. Despite being very challenging, having such video understanding is crucial for various vision applications such as autonomous driving, robot control, video editing, and augmented reality.
Video semantic segmentation has been regarded as one of the representative proxies for this ambitious goal. The community has proposed a large number of learning-based approaches under two main research directions of improving accuracy~\cite{fayyaz2016stfcn,gadde2017semantic,jin2017video,zhu2017flow,nilsson2018semantic,ding2020every} and efficiency~\cite{shelhamer2016clockwork,zhu2017deep,li2018low,jain2019accel,liu2020efficient}. While there has been a flurry of advances in model designs, significantly less effort has been made to the training objectives.
The main bottleneck of developing a useful supervision signal is the scarcity of video annotations available.
Most video segmentation benchmarks~\cite{cordts2016cityscapes,yu2020bdd100k} provide annotations for only a single frame per video clip, which limits these annotations still to an \textit{image} level. 

\begin{figure}[t]
\begin{center}
\begin{tabular}{@{}c@{}}
\includegraphics[width=1
\linewidth]{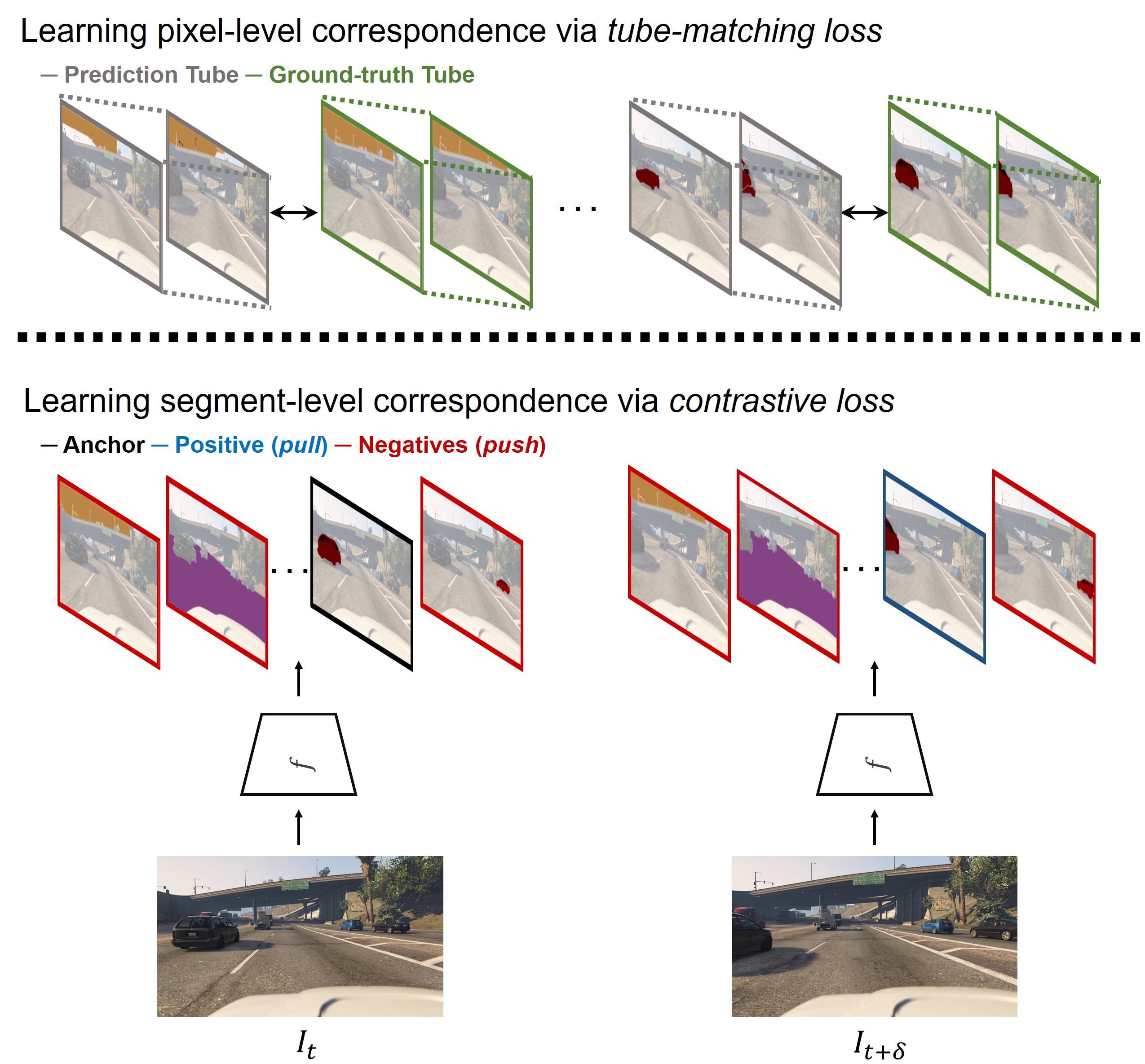} \\
\end{tabular}
\end{center}
\vspace{-6mm}
\caption{
\textbf{Overview of our approach.}
Temporal correspondence learning is fundamental for video understanding.
In this work, we extend this concept to every segment in a video.
Specifically, given a pair of frames, $I_{t}$ and $I_{t+\delta}$, a siamese model, $f$, learns to associate every segment at two different levels jointly: \textit{segment-level} via contrastive loss and \textit{pixel-level} via tube-matching loss.
}
\vspace{-3mm}
\label{fig:teaser}
\end{figure}

In line with the growing importance of the field and to meet the high demand for video labels, Kim~\etal~\cite{kim2020video} recently presented a new challenging dense video understanding problem called video panoptic segmentation (VPS) and presented a pair of video annotations: Cityscapes-VPS and VIPER datasets.
The problem unifies the existing video semantic segmentation~\cite{fayyaz2016stfcn,zhu2017flow,shelhamer2016clockwork} and video instance segmentation~\cite{yang2019video,bertasius2020classifying}. It aims at a simultaneous prediction of object classes, masks, instance id associations, and semantic segmentation for \textit{all} pixels in a video. The two VPS benchmarks provide real-world and simulation-based video datasets, opening up the possibilities to probe new video-specific training signals.

Identifying \textbf{\textit{temporal correspondence}} - ``what went where'' - in a video is a crucial requirement of robust visual reasoning in space and time.
Numerous methods for learning temporal correspondences, from pixel-level to object-level, have been developed so far, e.g., optical-flow estimation~\cite{liu2010sift,ilg2017flownet,sun2018pwc} and object tracking~\cite{bertinetto2016fully,tao2016siamese,valmadre2017end,wang2017dcfnet,wang2018learning}.
However, most of the previous methods aim at addressing a single-level of temporal correspondence at a time, and less attempt has been made to solve different levels of temporal correspondences jointly.
For the dense panoptic segmentation of a video, we argue that the model representations should support reasoning at various levels of temporal correspondences.
At the same time, this should be considered for every individual segment~\footnote{Throughout the paper, we use the term \textit{\textbf{segment}} to denote the region of both foreground \textit{things} and background \textit{stuffs} in the video.} in a  video. 
To this end, we train a video segmenter that simultaneously learns correspondences across frames at both segment-level and pixel-level (see~\figref{fig:teaser}).
In the following, we describe these two important views in turn:

\noindent\textit{\textbf{1) Segment-level correspondence learning:}}
Let's imagine we assign an id tag for each segment in a video. 
Naturally, we provide the same id tag for the same segment over time.
We formulate this concept as a graph matching problem. 
In practice, we construct the graph from video frames, where nodes encode segments in each frame and edges are affinities between them.
We then aim to learn features such that strong edges represent temporal correspondences.
This is achieved through the \textit{contrastive learning}~\cite{hadsell2006dimensionality}, which can encourage pairs of segment embedding to have strong edges if they are temporally associated or weak edges otherwise.
Also, as time goes, the segments in a video undergo dynamic appearance changes such as deformation, occlusion, and perspective distortion.
Therefore, without composing multiple handcrafted data augmentations~\cite{oord2018representation,bachman2019learning,he2020momentum,chen2020improved,chen2020simple,grill2020bootstrap,caron2020unsupervised}, our formulation of matching the same segment at temporally distant frames naturally leads to learning segment representations invariant in such visual distortions. 

\noindent\textit{\textbf{2) Pixel-level correspondence learning:}}
The optical flow provides dense pixel-level correspondences where each pixel in the current frame goes in the next frame.
The photometric loss is often used for learning this.
Recent works~\cite{ding2020every,liu2020efficient} generalize the photometric loss to the logit domain, i.e., segmentation output, with the motivation of predicting temporally consistent labels.
However, as warping loss often assumes independence between each pixel, higher-level correlations or structures over pixels are hardly modeled.
To alleviate this issue, we introduce ``tubes'' formed by linking the segment masks along the time axis.
The model then learns to minimize the mismatch between the prediction tubes and the ground-truth tubes, 
globally optimizing the entire chain of intermediate mask predictions.
This constraint allows the model to capture fine-grained changes in segments, e.g., shape, boundary, and motion tendencies over time.
Empirically, we show that the proposed mask tube matching loss performs better than the warping loss in learning an accurate segmentation model.

We aim to learn these two different levels of temporal correspondences jointly. 
To do so, we propose an efficient framework.
Specifically, we use a deep siamese model and train with a pair of frames, where a neighbor reference frame in a pair is randomly selected with a time gap relative to the current target frame.
Our model can encode strong temporal consistency into the features during training without using any heavy feature aggregation~\cite{zhu2017flow} or fusion~\cite{wang2019edvr} operations, resulting in an efficient yet strong video model.
Without bells and whistles, our model achieves new state-of-the-art results on Cityscapes-VPS and VIPER datasets while running in a fraction of time compared to the previous state-of-the-art approach~\cite{kim2020video}.
We summarize the contributions of this paper as follows.
\begin{enumerate}[topsep=0pt,itemsep=0pt]
\item We generalize the temporal correspondence learning to every segment in a video.
We present to learn coarse segment-level matching and fine pixel-level matching together.
We achieve this by designing two novel objective functions with an efficient learning framework.
\item We propose a new supervised\footnote{We effectively leverage the video label information.} contrastive learning method to learn the temporal correspondences in a video. In particular, we aim to maximize the mutual information between representations of temporally distant frames of the same segment.
\item We achieve new state-of-the-art on benchmarks, clearly demonstrating the effectiveness of our approach. We additionally provide extensive experimental analysis with ablation studies.
\end{enumerate}

\section{Related Works}
\paragraph{Video Segmentation}
Video segmentation aims to assign pixel-wise semantic labels to video frames.
As an important task for dynamic scene understanding, it has attracted attention increasingly from the research community.
Numerous approaches have been proposed in the literature, either focusing on improving the segmentation quality~\cite{fayyaz2016stfcn,gadde2017semantic,zhu2017flow,jin2017video,nilsson2018semantic,ding2020every} or accelerating the computation speed~\cite{shelhamer2016clockwork,zhu2017deep,li2018low,jain2019accel,liu2020efficient}.
With the growing importance of the field, a more fine-grained dynamic scene understanding task, video panoptic segmentation~\cite{kim2020video}, has been recently presented. It unifies semantic segmentation, instance segmentation, and multi-object tracking into a single coherent task.
Kim~\etal~\cite{kim2020video} designed a strong baseline model called VPSNet that consists of a shared backbone with multiple task-specific heads. 
While effective, the model enforces only weak temporal consistency by aggregating neighboring features using optical-flow~\cite{ilg2017flownet} and tracking foreground objects~\cite{yang2019video}. Thus, they missed leveraging intrinsic temporal information in a video fully.
Our model instead benefits from learning strong temporal consistency, from segment-level to pixel-level, through the proposed learning objectives during training.
With our learning strategy, we show that our per-frame inference model can even generate more accurate video segmentation results than the previous state-of-the-art method.

\begin{figure*}[t]
\begin{center}
\begin{tabular}{@{}c@{}}
\includegraphics[width=1
\linewidth]{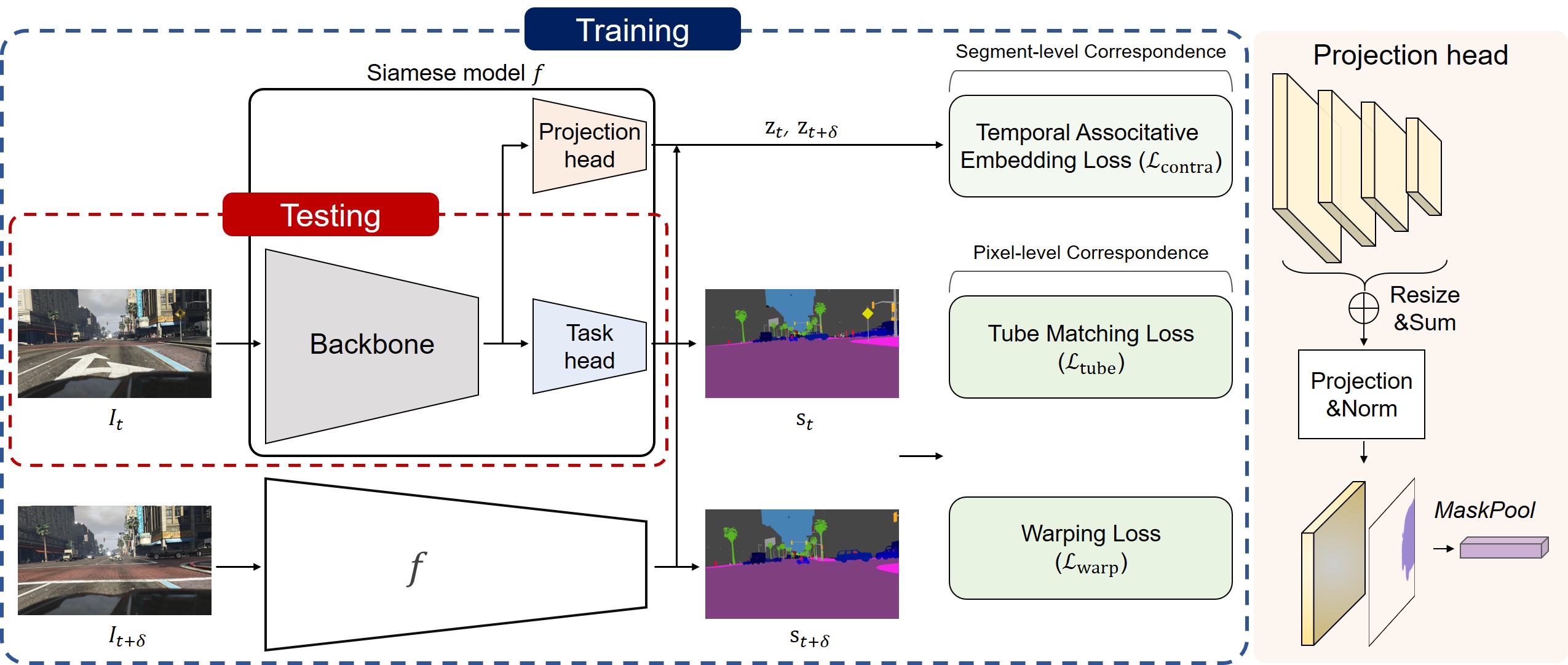} \\
\end{tabular}
\end{center}
\vspace{-7mm}
\caption{(\textbf{Left}) The overall pipeline of our joint learning framework. We utilize a deep siamese model and train coarse \textit{segment-level} matching and fine \textit{pixel-level} matching together. At inference time, we process video frames in a per-frame manner, without relying on heavy temporal feature aggregation or fusion. (\textbf{Right}) The detailed illustration of the projection head, which is required for the contrastive learning at train time only. The forward procedure of the projection head is: First, the FPN features are resized to the highest resolution level, and element-wise summed. Second, it is projected to a latent space via projection and normalization. Finally, the MaskPool operation is applied to obtain a segment-level feature embedding vector. \textit{ Best viewed in color.}}
\vspace{-3mm}
\label{fig:overview}
\end{figure*}

\paragraph{Temporal Correspondence Learning}
Temporal continuity provides a strong visual constancy that links disparate percepts into a smoothly-varying entity in the visual world. 
Learning temporal correspondence serves as a useful training signal in a video and it has been widely explored from pixel-wise to object-level such as optical flow estimation~\cite{liu2010sift,ilg2017flownet,sun2018pwc} and visual tracking~\cite{bertinetto2016fully,tao2016siamese,valmadre2017end,wang2017dcfnet,wang2018learning}.
However, there was less effort in learning different levels of temporal correspondences together, which are inherently related.
In this paper, we explore two different temporal stability constraints in a video and show the effectiveness of joint learning.
As large-scale video annotations are recently emerging through simulation~\cite{richter2016playing,richter2017playing,dosovitskiy2017carla,hurtado2020mopt} or automated label generation methods~\cite{porzi2020learning}, our work aims to develop commensurate video-specific learning strategies.
Another orthogonal direction is the self-supervised approach that constructs pretext tasks by exploiting temporal information in a video, e.g., cycle-consistency of time~\cite{wang2019learning,wang2019unsupervised,li2019joint,jabri2020space}.
Since the training data in the video domain is more likely to be a semi-supervised setting, we believe the progress in both directions, supervised and unsupervised learning in the video, should be made in parallel.

\paragraph{Contrastive Learning}
Recent resurgence in contrastive
learning has led to significant advances in a self-supervised learning paradigm~\cite{oord2018representation,bachman2019learning,he2020momentum,chen2020improved,chen2020simple,grill2020bootstrap,caron2020unsupervised}.
A fundamental idea of contrastive learning~\cite{hadsell2006dimensionality} is that similar examples should be grouped together and far from other clusters of related examples. It is achieved by using three key ingredients, an anchor, positive, and negative(s): pulling together an anchor and a positive sample in embedding space and pushing apart the anchor from many negative samples.
Data augmentations of the sample often construct a positive pair, and negative pairs are formed by the anchor and randomly chosen
samples from the minibatch.
Here, we aim to extend this current self-supervised contrastive learning paradigm from image to video. At the same time, from unsupervised to supervised by effectively leveraging the video label information.
Our work shares a similar objective with the recent study, SupCon~\cite{khosla2020supervised}, but we explore a different set of view definitions, architectures, and application settings. Also, we contribute a unique empirical investigation of supervised contrastive learning on the structured output space, i.e., segmentation.

\section{Method}

Video frames are temporally coherent in nature. 
A video panoptic segmentation model should faithfully exploit this temporal stability to seamlessly capture the whole segment's panoptic movement in a video. Otherwise, inconsistency in any class label and instance-id will result in low video quality of these panoptic segmentation sequences. 
This paper suggests learning the temporal correspondences in a video from two different perspectives jointly: \textit{segment} and \textit{pixel}.
In the following, we elaborate our approach in detail, including the network design (\secref{sec:network}) and objective functions (\secref{sec:loss}).
An overview of our method is shown in~\figref{fig:overview}.

\subsection{Network Architecture}
\label{sec:network}
In this paper, we design a deep siamese model, training on a pair of frames, where the neighbor reference frame is randomly sampled from a large range relative to the current target frame~\cite{zhu2017flow}. We then force the model to learn representations that can optimally associate the contents of the input frames.

Our model is fully convolutional and mainly consists of three parts: 
We first forward the target and reference images, $I_{t}$ and $I_{t+\delta}$, into a siamese network where each sub-network has the same structure with shared parameters. 
We then match the intermediate features (or predicted panoptic logits) of the two temporally distant images.
We finally compute the penalties to learn the temporal correspondences. 
All components are differentiable, allowing us to train the whole network end-to-end. 

\paragraph{Segmentation model.}
While not being sensitive to any specific
design of the segmentation model, we choose the previous state-of-the-art, VPSNet~\cite{kim2020video} as a baseline.
To avoid memory issues during the training, we use an efficient version, VPSNet-Track~\footnote{The heavy version, VPSNet-FuseTrack, cannot be trained with a pair of frames due to the GPU memory limit.}, without their memory-heavy Fuse module.
The model consists of a shared ResNet50-Feature Pyramid Network (FPN) backbone~\cite{lin2017feature} with multiple task-specific heads: Mask-RCNN~\cite{he2017mask}, deformable convolutions~\cite{dai2017deformable}, and MaskTrack~\cite{yang2019video} for instance segmentation, semantic segmentation, and tracking, respectively.

We propose to learn a projection head to map the feature pyramid representations to a latent embedding space, which will be used for our contrastive feature learning. The projection head is only deployed at training time and abandoned at inference. We provide the details of the projection head in the following.

\paragraph{Projection head.}
Recent works~\cite{chen2020simple,chen2020improved} introduce a small neural network so-called a projection head, e.g., MLP with one hidden layer, to map representations from the base encoder to a latent space where contrastive loss is applied.
In this context, we treat our backbone, i.e., ResNet50-FPN, as a base encoder and perform the projection on the FPN features. Our projection head mainly consists of two parts.
The first is a gathering operation where we merge multiple levels into a single, highest resolution level by resizing and elementwise adding the feature maps~\cite{pang2019libra}. Second, we then apply two $3\times3$ convolutions and a ReLU in between them, followed by the L2 normalization layer.
Finally, the resulting output feature is used for contrastive learning.
While various projection head implementations are possible, our simple design already achieves great performance.

\subsection{Objective Functions}
\label{sec:loss}
We exploit the temporal continuity in a video as a crucial supervisory signal, which endows strong temporal stability priors to the model.
To this end, we define the losses that guide the network to learn temporal correspondences from two different perspectives as
\begin{equation}
\begin{split}
\mathcal{L}_\mathrm{tc} =  \lambda_\mathrm{segment} \mathcal{L}_\mathrm{segment} + \lambda_\mathrm{pixel} \mathcal{L}_\mathrm{pixel},
\end{split}
\end{equation}
which consists of loss functions for segment-level correspondence learning $\mathcal{L}_\mathrm{segment}$ and pixel-level correspondence learning $\mathcal{L}_\mathrm{pixel}$.
In the following, we describe each term in detail.

\begin{figure*}[t]
\begin{center}
\begin{tabular}{@{}c@{}}
\includegraphics[width=1
\linewidth]{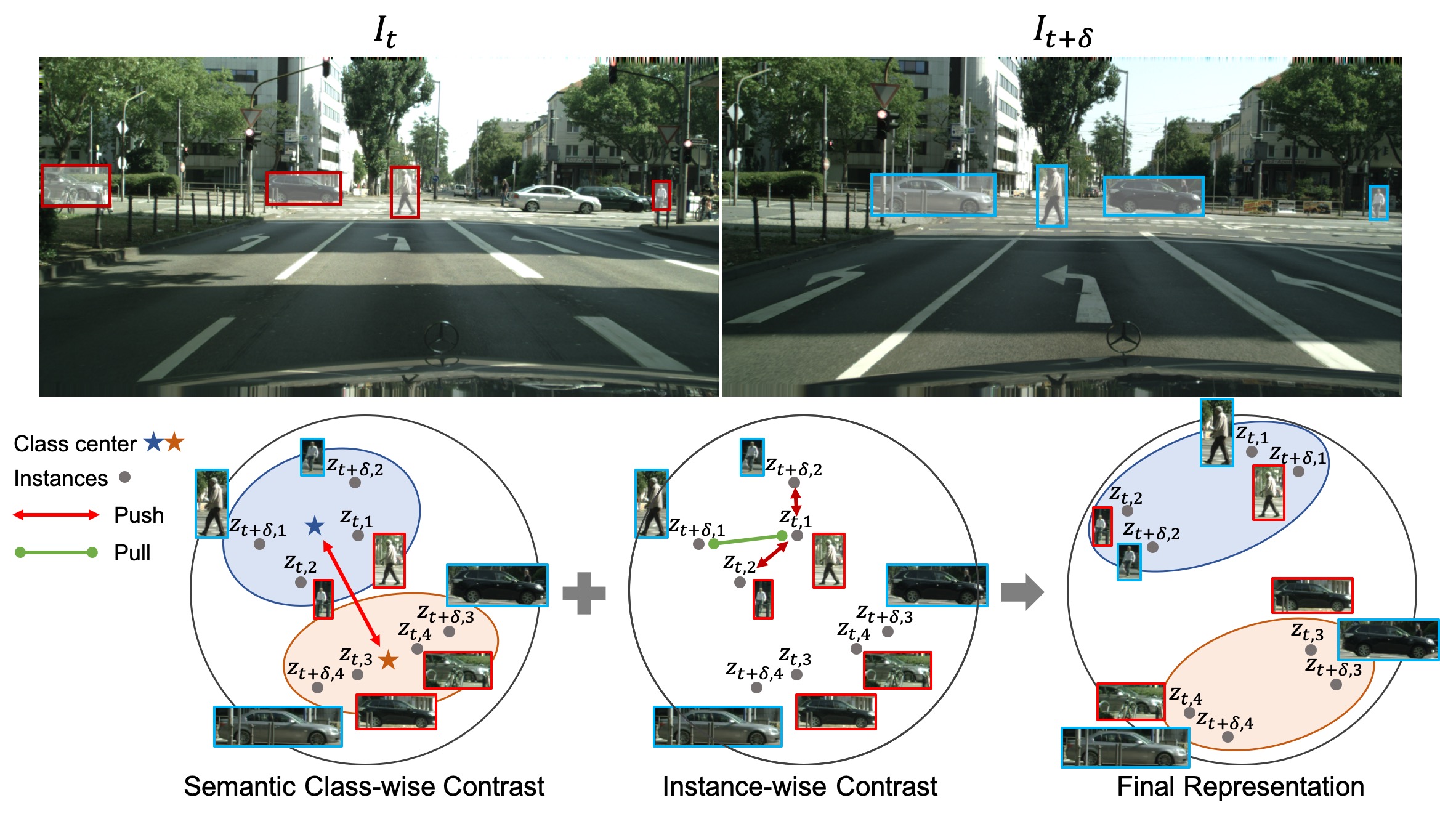} \\
\end{tabular}
\end{center}
\vspace{-9mm}
\caption{
We exploit two different contrastive forces for segment-level temporal correspondence learning: semantic class-wise contrast and instance-wise contrast.
The pull-push force at the semantic class-level enforces the model to learn \textit{\textbf{grouping}}.
In the meantime, the pull-push force at the instance-level enforces the model to learn the \textit{\textbf{instance discrimination}}.
These two contrasting forces are applied to link $I_{t}$ and $I_{t+\delta}$, learning \textit{\textbf{temporally invariant features}} as a consequence.
As learning proceeds, the feature embeddings desirable for video panoptic segmentation are finally learned.
For clarity, only the partial \textit{pull-push} operations are visualized.
}
\vspace{-3mm}
\label{fig:ta_concept}
\end{figure*}

\subsubsection{Segment-level Correspondence}
We formulate the learning of segment-level temporal correspondence as a graph matching problem.
The graph is constructed from a video, where nodes indicate segments in each frame and edges are affinities between them.
We aim to place large weights on a node pair that is temporally associated, i.e., finding a match.
The remaining issues are 1) how we represent the nodes in each frame and 2) how we encode the temporal correspondences by strong edges.

To tackle the first problem, we use a mask pooling operation, which is inspired by Lorenzo~\etal~\cite{porzi2020learning}.
They use it to enhance the RoI features for robust Re-ID.
By pooling the features under the instance mask, the background information is discarded, and only the necessary foreground information is left.
Unlike theirs, our main purpose is to have a representative embedding vector for every segment in a frame.
Therefore, we exploit the ground-truth \textit{panoptic} masks. Specifically, we use the \textit{things} and \textit{stuff} masks to spatially pool the projected feature map for all individual segments. We treat the obtained segment-level vectors as the nodes of a graph for each frame.

To address the second problem, our solution is to apply contrastive learning~\cite{khosla2020supervised}:
\textit{``pull'' the embeddings if they are temporally associated and ``push'' otherwise}.
Given a pair of temporally distant frames, $I_{t}$ and $I_{t+\delta}$, we only consider $N$ traceable segments.
Then the embeddings of the same segment in two different frames, $z_{t,i}$ and $z_{t+\delta,i}$\footnote{$i\in[1,N]$}, are regarded as a positive pair and the embeddings of other 2($N$-1) segments, e.g., $z_{t,j}$\footnote{$j\neq i$}, as negatives.
For a segment $i$, the loss function for a positive pair of examples $(t, t+\delta)$ is formulated as
\begin{equation}
\begin{split}
l^{i}_{t,t+\delta} = -\mathrm{log}\frac{\mathrm{exp(sim(\mathnormal{z_{t,i},z_{t+\delta,i}})/\tau)}}{\sum_{*} \sum_{k}\mathbbm{1}_{[k \neq i \land * \neq t]}\mathrm{exp(sim(\mathnormal{z_{t,i},z_{*,k}})/\tau)}}\\
\text{s.t.} \quad *\in\{t, t+\delta\},
\end{split}
\label{eqn:contra}
\end{equation}
where $\mathbbm{1}_{[k \neq i \land * \neq t]} \in \{0,1\}$ is an indicator function evaluating to 1 iff $k \neq i$ and $* \neq t$. The $\tau$ denotes a temperature parameter, which is set to 0.5. The final contrastive loss is computed across all positive pairs of segments as
\begin{equation}
\begin{split}
\mathcal{L}_{contra} = \frac{1}{2N}\sum\limits_{i=1}^N l^{i}_{t,t+\delta} + l^{i}_{t+\delta,t}\\
\end{split}
\label{eqn:contra_full}
\end{equation}

Finally, we found that how we contrast the embedding is also important for the performance. For example, if we only contrast the segment embeddings computed from the semantic segmentation label, the different instances in the same semantic class are not distinguished, which degrades the instance tracking accuracy.
On the other hand, using only the \textit{thing} instance masks leads to low classification accuracy as different segment embeddings of the same semantic pushed too far apart.
Also, the background regions are not considered in learning the temporal association.
To resolve these issues, our key idea is to construct two different graphs using both semantic and instance segmentation labels and apply the contrastive learning simultaneously but independently (see~\figref{fig:ta_concept}). Our formulation enables class-aware instance contrasting. 
As training proceeds, the features suitable for panoptic segmentation are learned. 
The loss for the segment-level correspondence learning is defined as
$\mathcal{L}_\mathrm{segment} = \mathcal{L}_\mathrm{contra\_sem} + \mathcal{L}_\mathrm{contra\_inst}$,
where $\mathcal{L}_\mathrm{contra\_sem}$ and $\mathcal{L}_\mathrm{contra\_inst}$ denote the contrastive loss (\eqnref{eqn:contra_full}) computed from the graphs using semantic segmentation and instance segmentation annotations, respectively.

\subsubsection{Pixel-level Correspondence}
A dense temporal correspondence learning is often achieved by using a photometric loss.
The loss is mainly used for flow estimation~\cite{ilg2017flownet,sun2018pwc} or temporal stability of output frames~\cite{lai2018learning,kim2019deep,park2019preserving}. 
It is calculated by warping the current frame to the next frame through the optical flow and computing the warping error.
Recent studies~\cite{ding2020every,liu2020efficient} generalize the photometric loss to the logit space, i.e., segmentation output, aiming at learning a segmentation model that predicts temporally consistent labels.
However, the photometric loss assumes pixel independence and thus is challenging to model complex correlations over pixels.
This may overlook learning important higher-level temporal structures in a video, such as a segment shape changes through time.

To this end, we introduce a concept of \textit{tube} that is constructed by connecting the segment masks along the time-axis.
The model attempts to globally minimize the mismatch between the prediction tubes and the ground-truth tubes during training.
The proposed tube matching loss is applied pixel-wise while allowing the model to capture important temporal structures over the pixels.
We show that it outperforms the warping loss in learning an accurate segmentation model.
However, we find that warping loss can complement the tube-matching loss, and the best performance can be achieved when using them together (see~\tabfref{tab:tc_anal}).
Therefore, we propose to use them jointly to achieve strong pixel-level temporal correspondence learning.
In the following, we detail how we define the final loss function for learning the pixel-level correspondences.

\paragraph{Warping loss.} Given a pair of video frames $I_{t}$ and $I_{t+\delta}$, we first feed them into the flow estimation network, e.g., FlowNet~\cite{ilg2017flownet}, and obtain the optical flow of each pixel from the current frame to the next frame, $F_{t\Rightarrow t+\delta}$. To avoid penalizing the pixels in the invisible regions, we estimate the occlusion map as  $O_{t+\delta \Rightarrow t}=\exp(-\alpha||I_{t}-\hat{I}_{t+\delta}||_2)$, where $\hat{I}_{t+\delta}$ is the frame $I_{t+\delta}$ warped by the flow $F_{t\Rightarrow t+\delta}$ and $\alpha$ is set to 50.
By the nature of the convolutional neural network, the output logit is spatially aligned with the input frame. 
We thus directly use the downsampled version\footnote{The spatial size of the logit is often smaller than original input resolution, e.g., 1/4 or 1/8.} of the same flow $(F')$ and occlusion map $(O')$ to constrain the logits.  The panoptic logits $s_{t}$ and $s_{t+\delta}$ are acquired by feeding the input frames into the segmentation model.
The warping loss is defined as
\begin{equation}
     \begin{split}
     \mathcal{L}_{warp}
     = \sum_{x,y} O'^{x,y}_{t \Rightarrow t+\delta} \left \| s^{x,y}_{t} - \hat{s}^{x,y}_{t+\delta} \right\|_{2},\\
     \end{split}
 \end{equation}
where $s^{x,y}$ is the logit vector at location $(x,y)$ and $\hat{s}_{t+\delta}$ is the logit $s_{t+\delta}$ warped by the flow $F'_{t\Rightarrow t+\delta}$.

\paragraph{Tube-matching loss.} Given a pair of panoptic logits $s_{t}$ and $s_{t+\delta}$, we first apply softmax layer along the the channel-axis. 
Then, each channel of the panoptic logit will encode specific segment masks, and the values will range between 0 and 1.
We temporally concatenate the same segment masks in two different panoptic logits (see~\figref{fig:teaser}).
Formally, for an segment $i$, its tube, $T_{i}$, can be constructed as
\begin{equation}
    \begin{split}
    \dot{s} = Softmax(s) \\
    T_{i} = [\dot{s}_{t,i}, \dot{s}_{t+\delta,i}] \\ 
    \end{split}
\end{equation}
where $\dot{s}_{t,i}$ and $\dot{s}_{t+\delta,i}$ denote the mask of segment $i$ in the frame $I_{t}$ and $I_{t+\delta}$, respectively.
For the tube matching, we adopt the dice coefficient~\cite{dice1945measures}, $D$, which is insensitive to the number of foreground and background pixels and can have balanced importance on every segment regardless of their size.
It is defined as,
\begin{equation}
    \begin{split}
    D(p,q) = \frac{2\sum (p \cdot q)}{\sum p^2 + \sum q^2}.
    \end{split}
\end{equation}
As the dice coefficient has a range from 0 to 1, the tube matching loss function is calculated as
\begin{equation}
    \begin{split}
    \mathcal{L}_{tube}
    = \frac{1}{N} \sum_{i} 1 - D(T_{i}, \tilde{T}_{i}), 
    \end{split}
\end{equation}
where $\tilde{T}$ is the corresponding binary ground-truth mask tube. N is the total number of the traceable segments. 

The loss for the pixel-level correspondence learning is composed of these two losses as
$\mathcal{L}_{pixel}=\mathcal{L}_\mathrm{warp}+\mathcal{L}_\mathrm{tube}$.

\begin{table*}[t]
\centering
\resizebox{\textwidth}{!}{%
\setlength{\tabcolsep}{2.3pt}
\begin{adjustbox}{max width=\textwidth}
\begin{tabular}[b]{ l|ccc|c|c|c|c|c|c}
        \hline
        {\textbf{VPSNet-variants}}                                      &Segment &\multicolumn{2}{c|}{Pixel} &\multicolumn{4}{c|}{Temporal window size} & \multirow{2}{*}{VPQ} & \multirow{2}{*}{FPS} \\
        \cline{5-8} {on \textbf{VIPER}} & \textbf{$\mathcal{L}_\mathrm{segment}$} & \textbf{$\mathcal{L}_\mathrm{warp}$} &  \textbf{$\mathcal{L}_\mathrm{tube}$}  & k = 0 & k = 5 & k = 10 & k = 15 &  \\
        \hline
        Track  &		 
        & 		&    & 
        48.1 / 38.0 / 57.1 & 
        49.3 / 45.6/ 53.7  & 
        45.9 / 37.9 / 52.7 & 
        43.2 / 33.6 / 51.6 & 
        46.6 / 38.8 / 53.8 &
        5.1\\
        FuseTrack~\cite{kim2020video}   & 
        &      &    & 
        49.8 / 40.3 / 57.7 &
        51.6 / 49.0 / 53.8 &
        47.2 / 40.4 / 52.8 &                        
        45.1 / 36.5 / 52.3 &
        48.4 / 41.6 / 53.2 &
        1.6\\
        SiamTrack-\textit{Base} \quad                                  &       
        &      &    & 
        49.5 / 39.1 / 58.1 
        & 50.8 / 46.6 / 54.0 
        & 46.7 / 38.8 / 53.0 
        & 44.5 / 34.5 / 52.4                        
        & 47.9 / 39.8 / 54.3  &
        5.1\\
        SiamTrack \quad                                  & 
        \checkmark &  &  & 
        50.7 / 41.7 / 58.2 &  			
        52.6 / 50.8 / 54.2 & 
        48.4 / 42.9 / 53.3 & 
        46.5 / 39.3 / 52.8 &                        
        49.6 / 43.7 / 54.6 &
        5.1\\
        SiamTrack \quad                                  & 
        & \checkmark &  & 
        50.3 / 40.4 / 58.6 & 
        51.3 / 47.6 / 54.3 &  			
        47.3 / 40.0 / 53.3 & 
        45.4 / 36.4 / 52.9 & 
        48.6 / 41.1 / 54.8  &
        5.1\\
        SiamTrack \quad                                   &  
        &  &\checkmark  & 
        50.7 / 41.0 / 58.6 & 
        52.4 / 49.6 / 54.6 & 
        48.5 / 42.5 / 53.5 &                       
        46.7 / 39.3 / 52.8 &      		
        49.6 / 43.1 / 54.9  &
        5.1\\
        SiamTrack-\textit{Ours} \quad                           & 
        \checkmark &  \checkmark& \checkmark     & 
        \textbf{51.1} / 42.3 / 58.5 & 
        \textbf{53.4} / 51.9 / 54.6 & 
        \textbf{49.2} / 44.1 / 53.5 & 
        \textbf{47.2} / 40.3 / 52.9 & 
        \textbf{50.2} / 44.7 / 55.0 &
        5.1\\
        \hline
\end{tabular}
\end{adjustbox}
}
\vspace{-6mm}
\caption{Video panoptic segmentation results on VIPER validation set. Each cell contains VPQ / VPQ\textsuperscript{Th} / VPQ\textsuperscript{St} scores.}
\label{tab:viper_vpq}
\end{table*}

\begin{table*}[]
\centering
\resizebox{\textwidth}{!}{%
\begin{adjustbox}{max width=\textwidth}
\begin{tabular}{l|c|c|c|c|c|c}
\hline
{\textbf{VPSNet-variants}} &\multicolumn{4}{c|}{Temporal window size} 
                & \multirow{2}{*}{VPQ} & \multirow{2}{*}{FPS} \\
\cline{2-5} {on \textbf{Cityscapes-VPS \textit{val.}}} & k = 0 & k = 5 & k = 10 & k = 15 &  \\
\hline
Track  & 
			63.1 / 56.4 / 68.0 & 
            56.1 / 44.1 / 64.9 & 
            53.1 / 39.0 / 63.4 &
            51.3 / 35.4 / 62.9 &   
            55.9 / 43.7 / 64.8 & 4.5\\

FuseTrack~\cite{kim2020video}  \quad   & 
            64.5 / 58.1 / 69.1 & 
            57.4 / 45.2 / 66.4 &
            54.1 / 39.5 / 64.7 &   
            52.2 / 36.0 / 64.0 & 
            57.2 / 44.7 / 66.6 & 1.3\\
SiamTrack-\textit{Base}
\quad   & 
            63.2 / 57.0 / 67.7 & 
            56.2 / 42.6 / 66.3 &
            53.2 / 38.3 / 65.2 &   
            51.5 / 33.7 / 64.5 & 
            56.0 / 42.9 / 65.8 & 4.5 \\
SiamTrack-\textit{Ours} 
\quad   & 
            \textbf{64.6} / 58.3 / 69.1 & 
            \textbf{57.6} / 45.6 / 66.6 &
            \textbf{54.2} / 39.2 / 65.2 &   
            \textbf{52.7} / 36.7 / 64.6 & 
            \textbf{57.3} / 44.7 / 66.4 & 4.5 \\ 
\hline
\end{tabular}
\end{adjustbox}
}
\\
\vspace{2mm}
\centering
\resizebox{\textwidth}{!}{%
\begin{adjustbox}{max width=\textwidth}
\begin{tabular}{l|c|c|c|c|c|c}
\hline
{\textbf{VPSNet-variants}} &\multicolumn{4}{c|}{Temporal window size} 
                & \multirow{2}{*}{VPQ} & \multirow{2}{*}{FPS} \\
\cline{2-5} {on \textbf{Cityscapes-VPS \textit{test}}} & k = 0 & k = 5 & k = 10 & k = 15 &  \\
\hline
Track  & 
			63.1 / 58.0 / 66.4 & 
            56.8 / 45.7 / 63.9 & 
            53.6 / 40.3 / 62.0 &
            51.5 / 35.9 / 61.5 &     
            56.3 / 45.0 / 63.4 & 4.5\\
FuseTrack~\cite{kim2020video} \quad  & 
			\textbf{64.2} / 59.0 / 67.7 & 
            57.9 / 46.5 / 65.1 & 
            54.8 / 41.1 / 63.4 &
            52.6 / 36.5 / 62.9 &   
            57.4 / 45.8 / 64.8 & 1.3\\
SiamTrack-\textit{Base} \quad  & 
			63.2 / 58.6 / 66.1 & 
            57.1 / 46.9 / 63.6 & 
            54.0 / 41.8 / 62.6 &
            52.1 / 37.9 / 61.5 &     
            56.6 / 46.4 / 63.4 & 4.5\\
SiamTrack-\textit{Ours} \quad & 
            63.8 / 59.4 / 66.6 & 
            \textbf{58.2} / 47.2 / 65.9 &
            \textbf{56.0} / 43.2 / 64.4 &   
            \textbf{54.7} / 40.2 / 63.2 & 
            \textbf{57.8} / 47.5 / 65.0 & 4.5\\

\hline
\end{tabular}
\end{adjustbox}

}
\vspace{-2mm}
\caption{Video panoptic segmentation results on Cityscapes-VPS validation (top) and test (bottom) set with our VPSNet-variants. Each cell contains VPQ / VPQ\textsuperscript{Th} / VPQ\textsuperscript{St} scores.}
\vspace{-3mm}
\label{tab:cityvps_vpq}
\end{table*}

\section{Experiments}

This section presents the experimental results on the two video-level datasets, \textit{VIPER}, and \textit{Cityscapes-VPS}. 
We adopt the evaluation metrics of VPQ~\cite{kim2020video}, which reflect both image-level prediction and cross-frame association performance.

\subsection{Data} 
We use recently presented video panoptic segmentation datasets: VIPER and Cityscapes-VPS.

\begin{itemize}[topsep=0.5pt,itemsep=0.5pt]
\item \textbf{VIPER}: 
 We follow the public train / val split. For evaluation, we include all the validation videos from \textit{day} scenario and use the first 50 frames of each video: setting a total of 600 frames.
\item \textbf{Cityscapes-VPS}:
 We follow the public train / val / test split\footnote{400 training, 50 validation, and 50 test videos}. Each video consists of 30 consecutive frames, with every 5 frames paired with the ground-truth annotations. All 30 frames are predicted for each video, and only the 6 frames with the ground-truth are evaluated.
\end{itemize}

\begin{figure*}[t]
\begin{center}
\begin{tabular}{@{}c@{}}
\includegraphics[width=1\linewidth]{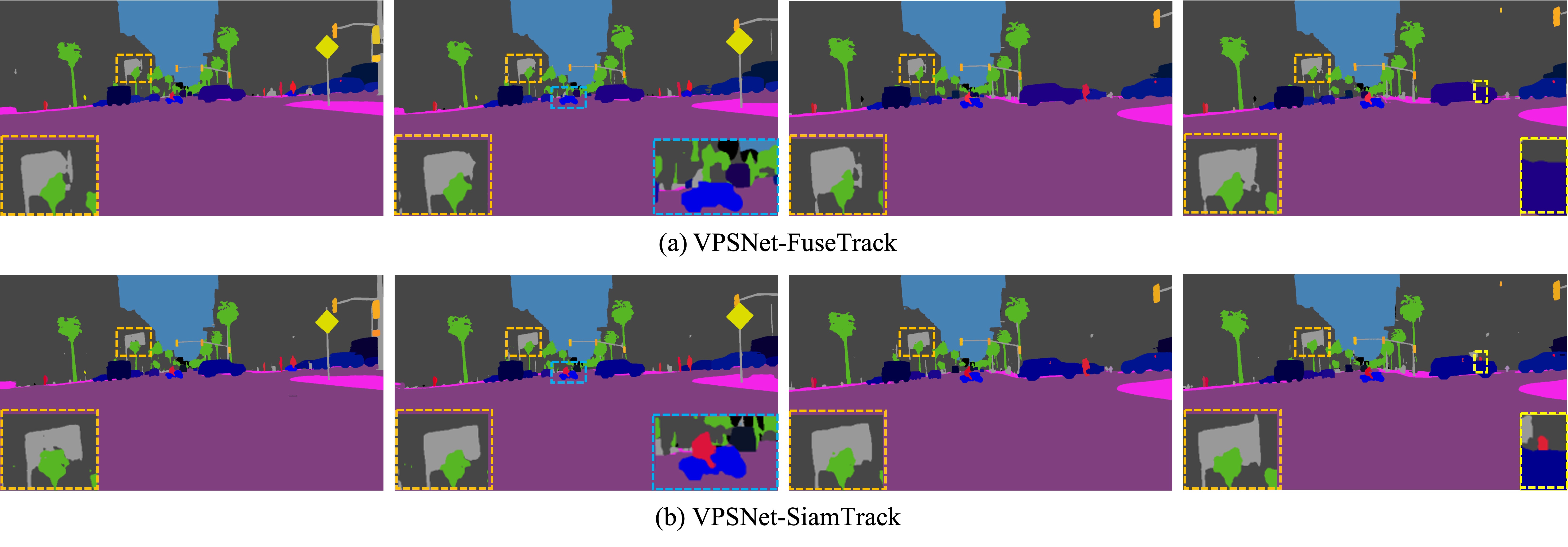} \\
\end{tabular}
\end{center}
\vspace{-9mm}
\caption{
The qualitative results on VIPER. Frames are sampled uniformly. The insets show enlarged views of the dotted boxes.
}
\vspace{-3mm}
\label{fig:qual}
\end{figure*}

\subsection{Quantitative Results} 

We consider VIPER as a primary evaluation dataset, based on its high quantity and quality of the annotation.
We mostly conduct experiments with this benchmark.

\paragraph{Baselines.} 
We compare our approach to the state-of-the-art VPSNet~\cite{kim2020video}.
The followings are the detailed information of the baseline models used in experiments.
\begin{itemize}[topsep=0.5pt,itemsep=0.5pt]
\item \textbf{VPSNet-Track}: 
A basic video panoptic segmentation model, which adds tracking head~\cite{yang2019video} on top of the image panoptic segmentation model, UPSNet~\cite{xiong2019upsnet}.
\item \textbf{VPSNet-FuseTrack}: 
An advanced model, which additionally supports temporal feature aggregation~\cite{zhu2017flow} and attentional space-time feature fusion~\cite{wang2019edvr}. This model holds current state-of-the-art VPQ scores.
\item \textbf{VPSNet-SiamTrack}: 
We use VPSNet-Track as a base segmentation model and train it with a pair of temporally distant frames.
As there is no difference in model designs, we can see the performance improvement of \textbf{SiamTrack} over the \textbf{Track} totally comes from the enhanced model representations after the training.
\end{itemize}

\subsubsection{Main Results} 
\paragraph{VIPER.} 
First, we demonstrate the impact of the proposed learning objectives.
The results are summarized in~\tabfref{tab:viper_vpq}.
The baseline is a siamese model without using any temporal correspondence learning at train time.
Upon this baseline, we apply our proposals.
As shown in the table, we find that all the loss functions improve the baseline with significant margins, showing their effectiveness.
Specifically, the VPQ scores increased by $1.7\%$, $0.7\%$, and $1.7\%$ for $\mathcal{L}_\mathrm{contra}$, $\mathcal{L}_\mathrm{warp}$, and $\mathcal{L}_\mathrm{tube}$, respectively.
Note that our learning objectives provide the model to perform well on the more challenging regime, i.e., the temporal window size of $k>=5$.
The best VPQ scores were achieved when the two objectives, segment-level and pixel-level temporal correspondence, are enforced together. 
This result implies that correspondence learning at both segment-level and pixel-level synergize each other to learn more discriminative video features.
Compared to the state-of-the-art, VPSNet-FuseTrack, our final model achieves much higher VPQ scores (48.4 VPQ vs. \textbf{50.2 VPQ}).
Moreover, as our model does not rely on temporal aggregation~\cite{zhu2017flow} and fusion~\cite{wang2019edvr} strategies, we enjoy faster inference speed (1.6FPS vs. \textbf{5.1FPS}) and lower memory usage (5.5G vs. \textbf{3.7G}) than the VPSNet-FuseTrack.
This shows great potential of our learning framework for practical applications on high-speed and low throughput scenarios.

\paragraph{Cityscapes-VPS.}
We also benchmark our model on Cityscapes-VPS \textit{val} and \textit{test}. 
The results are summarized in~\tabref{tab:cityvps_vpq}.
We observed a consistent tendency, where our full VPSNet-SiamTrack model achieves +1.3\%VPQ and +1.2\%VPQ higher than the baseline for \textit{val} and \textit{test}, respectively. 
Also, it achieves better performance compared to the state-of-the-art, while running in much faster inference speed ($3\times$) and using lower memory usage (5.5G vs. \textbf{3.7G}).
We see the relatively lower performance improvement than the VIPER mainly comes from the difference in annotation quality~\footnote{We found several critical annotation errors in the
Cityscapes-VPS; incorrect track-ids, wrong class labels,
and low-quality masks. They may lead to noisy training
signal (e.g., wrong temporal association of the segments or
inaccurate boundaries and shapes).}.

\paragraph{Analysis on $\mathcal{L}_\mathrm{segment}$.} 
The model learns coarse segment-level temporal correspondence by matching their embeddings across frames.
The temporal association is achieved by contrastive learning, and for successful learning, it is essential to set positive and negatives properly~\cite{he2020momentum,chen2020improved,chen2020simple}.
Here, we consider the same segment's embeddings in the different frames as positive and others as negatives.
During learning, our approach exploits both the semantic and instance segmentation labels.
The results are summarized in~\tabref{tab:ta_ablation}.
The instance-wise contrast ($\mathcal{L}_\mathrm{contra\_inst}$) enables the learning of visual features suitable for tracking.
However, as learning is unaware of the semantic-class, the instance features in the same class may vary significantly (49.6VPQ $\rightarrow$ 48.4VPQ).
Moreover, the background is ignored.
On the other hand, the semantic class-wise contrast ($\mathcal{L}_\mathrm{contra\_sem}$) enforces the model to learn grouping the same semantic features.
However, without instance-level contrasting, it is easily trapped into degeneracy (49.6VPQ $\rightarrow$ 47.6VPQ).
To this end, we present to use both the instance-level and semantic class-level contrasting together. 
The strong empirical results in~\tabfref{tab:ta_ablation} confirms that the features suitable for video panoptic segmentation is learned.

\paragraph{Analysis on $\mathcal{L}_\mathrm{pixel}$.}
The fine pixel-level correspondence is learned with the combination of warping loss ($\mathcal{L}_\mathrm{warp}$) and tube matching loss ($\mathcal{L}_\mathrm{tube}$).
We evaluate their impacts on both segmentation quality and output temporal smoothness using VPQ and TC~\cite{liu2020efficient}, respectively.
The results are summarized in~\tabfref{tab:tc_anal}.
As shown in the table, we can clearly see that the tube matching loss outperforms warping loss significantly, both on the segmentation quality (VPQ) and the smoothness of output videos (TC).
This backs our claim that modeling higher-level temporal structures over the pixels is more beneficial.
Also, we observe that employing both the tube matching loss and the warping loss provides the best VPQ and TC scores. 
This implies that both are complementary in constructing an accurate and temporally stable VPS model, showing the great advantage of our final loss form.

\paragraph{Analysis on time gap $\delta$.}
We also study the impact of the time gap, $\delta$, which allows the siamese model to exploit temporally distant frames during training. 
We conduct experiments with three different values of 5, 10, 15. On VIPER, each provided a final VPQ score of 49.8, 50.2, and 49.9, respectively. We thus use  $\delta \in \{-10,10\}$ .

\begin{table}[]
\centering
\resizebox{0.9\linewidth}{!}{%
\begin{tabular}[b]{ c|cc|ccc}
        \hline
        \multirow{2}{*}{Loss} & \multicolumn{2}{c|}{Segment-lvl matching} & \multirow{2}{*}{VPQ} & \multirow{2}{*}{VPQ\textsuperscript{Th}}  & \multirow{2}{*}{VPQ\textsuperscript{St}} \\
                              & $\mathcal{L}_\mathrm{contra\_inst}$            & $\mathcal{L}_\mathrm{contra\_sem}$             &      &      & \\
        \hline
                                           &       	    &             & 47.9          & 39.8    & 54.3 \\
                                           &\checkmark	&             & 48.4          & 40.6    & 54.2 \\
                                           &            & \checkmark  & 47.6          & 39.6    & 54.1 \\
        $\mathcal{L}_\mathrm{segment}$       &\checkmark  & \checkmark  & \textbf{49.6} & \textbf{43.7} & \textbf{54.6} \\
        \hline
\end{tabular}
} 
\vspace{-2mm}
\caption{Ablation studies on segment-level correspondence loss using VIPER.}
\label{tab:ta_ablation}
\vspace{-2mm}
\end{table}

\begin{table}[]
\centering
\resizebox{0.9\linewidth}{!}{%
\begin{tabular}[b]{ c|cc|cccc}
        \hline
        \multirow{2}{*}{Loss} & \multicolumn{2}{c|}{Pixel-lvl matching} & \multirow{2}{*}{VPQ} & \multirow{2}{*}{VPQ\textsuperscript{Th}}  & \multirow{2}{*}{VPQ\textsuperscript{St}} &  \multirow{2}{*}{TC~\cite{liu2020efficient}} \\
                              & $\mathcal{L}_\mathrm{warp}$            & $\mathcal{L}_\mathrm{tube}$             &      &      &  &\\
        \hline
                                           &       	    &             & 47.9          & 39.8    & 54.3  & 78.3\\
                                           &\checkmark	&             & 48.6          & 41.1    & 54.8  & 79.3\\
                                           &            & \checkmark  & 49.6          & 43.1    & 54.9  & 80.8\\
        $\mathcal{L}_\mathrm{pixel}$       &\checkmark  & \checkmark  & \textbf{49.9} & \textbf{43.8} & \textbf{55.0} &\textbf{82.2}\\
        \hline
\end{tabular}
} 
\vspace{-2mm}
\caption{Ablation studies on pixel-level correspondence loss using VIPER.}
\label{tab:tc_anal}
\vspace{-2mm}
\end{table}

\subsection{Qualitative Results}
In~\figref{fig:qual}, we compare our final model VPSNet-SiamTrack with VPSNet-FuseTrack~\cite{kim2020video} qualitatively.
The blue and yellow dotted boxes in the Figure 4 show the segment-level smoothness (i.e., the rider and pedestrian are continuously tracked and segmented over time). 
Meanwhile, the orange dotted box demonstrates the pixel-level smoothness (i.e., the shape and the boundary of the flag are temporally consistent). 
These two different temporal stability priors are encoded into the feature representations during the training.

\section{Conclusion}

In this paper, we generalize the temporal correspondence learning to every segment in a video.
To learn their temporal association, we present two novel objective functions with an efficient learning framework.
With our learning strategy, a per-frame inference model can outperform previous state-of-the-art while running in a fraction of time.
Last but not least, this is the first time to propose a supervised contrastive learning method for learning dense temporal associations. 
Based on our proposals and their empirical results, we hope more effective video-specific supervisory signals being presented in the future.

\paragraph{Acknowledgements}
This work was supported in part by Samsung Electronics Co., Ltd (G01200447). 
Sanghyun Woo and Dahun Kim are supported by Microsoft Research Asia Fellowship.

\clearpage

{\small
\bibliographystyle{ieee_fullname}
\bibliography{egbib}
}

\end{document}